# Encoding of phonology in a recurrent neural model of grounded speech

**Afra Alishahi**
Tilburg University
a.alishahi@uvt.nl

**Marie Barking**
Tilburg University
m.barking@uvt.nl

**Grzegorz Chrupała**
Tilburg University
g.chrupala@uvt.nl

## Abstract

We study the representation and encoding of phonemes in a recurrent neural network model of grounded speech. We use a model which processes images and their spoken descriptions, and projects the visual and auditory representations into the same semantic space. We perform a number of analyses on how information about individual phonemes is encoded in the MFCC features extracted from the speech signal, and the activations of the layers of the model. Via experiments with phoneme decoding and phoneme discrimination we show that phoneme representations are most salient in the lower layers of the model, where low-level signals are processed at a fine-grained level, although a large amount of phonological information is retain at the top recurrent layer. We further find out that the attention mechanism following the top recurrent layer significantly attenuates encoding of phonology and makes the utterance embeddings much more invariant to synonymy. Moreover, a hierarchical clustering of phoneme representations learned by the network shows an organizational structure of phonemes similar to those proposed in linguistics.

## 1 Introduction

Spoken language is a universal human means of communication. As such, its acquisition and representation in the brain is an essential topic in the study of the cognition of our species. In the field of neuroscience there has been a long-standing interest in the understanding of neural representations of linguistic input in human brains, most commonly via the analysis of neuro-imaging data of participants exposed to simplified, highly controlled inputs. More recently, naturalistic data has been used and patterns in the brain have been correlated with patterns in the input (e.g. Wehbe et al., 2014; Khalighinejad et al., 2017).

This type of approach is relevant also when the goal is the understanding of the dynamics in complex neural network models of speech understanding. Firstly because similar techniques are often applicable, but more importantly because the knowledge of how the workings of artificial and biological neural networks are similar or different is valuable for the general enterprise of cognitive science.

Recent studies have implemented models which learn to understand speech in a weakly and indirectly supervised fashion from correlated audio and visual signal: Harwath et al. (2016); Harwath and Glass (2017); Chrupała et al. (2017a). This is a departure from typical Automatic Speech Recognition (ASR) systems which rely on large amounts of transcribed speech, and these recent models come closer to the way humans acquire language in a grounded setting. It is thus especially interesting to investigate to what extent the traditional levels of linguistic analysis such as phonology, morphology, syntax and semantics are encoded in the activations of the hidden layers of these models. There are a small number of studies which focus on the syntax and/or semantics in the context of neural models of written language (e.g. Elman, 1991; Frank et al., 2013; Kádár et al., 2016; Li et al., 2016a; Adi et al., 2016; Li et al., 2016b; Linzen et al., 2016). Taking it a step further, Gelderloos and Chrupała (2016) and Chrupała et al. (2017a) investigate the levels of representations in models which learn language from phonetic transcriptions and from the speech signal, respectively. Neither of these tackles the representation of phonology in any great depth. Instead they work with relatively coarse-grained distinctions between form and meaning.

In the current work we use controlled synthetic stimuli, as well as alignment between the audio signal and phonetic transcription of spoken utterances to extract phoneme representation vectors based on the activations on the hidden layers of a model of grounded speech perception. We use these representations to carry out analyses of the representation of phonemes at a fine-grained level. In a series of experiments, we show that the lower layers of the model encode accurate representations of the phonemes which can be used in phoneme identification and classification with high accuracy. We further investigate how the phoneme inventory is organised in the activation space of the model. Finally, we tackle the general issue of the representation of phonological form versus meaning with a controlled task of synonym discrimination.

Our results show that the bottom layers in the multi-layer recurrent neural network learn invariances which enable it to encode phonemes independently of co-articulatory context, and that they represent phonemic categories closely matching usual classifications from linguistics. Phonological form becomes harder to detect in higher layers of the network, which increasingly focus on representing meaning over form, but encoding of phonology persists to a significant degree up to the top recurrent layer.

We make the data and open-source code to reproduce our results publicly available at github.com/gchrupala/encoding-of-phonology.

## 2 Related Work

Research on encoding of phonology has been carried out from a psycholinguistics as well as computational modeling perspectives. Below we review both types of work.

### 2.1 Phoneme perception

Co-articulation and interspeaker variability make it impossible to define unique acoustic patterns for each phoneme. In an early experiment, Liberman et al. (1967) analyzed the acoustic properties of the /d/ sound in the two syllables /di/ and /du/. They found that while humans easily noticed differences between the two instances when /d/ was played in isolation, they perceived the /d/ as being the same when listening to the complete syllables. This phenomenon is often referred to as categorical perception: acoustically different stimuli are perceived as the same. In another experiment Lisker and Abramson (1967) used the two syllables /ba/ and /pa/ which only differ in their *voice onset time* (VOT), and created a continuum moving from syllables with short VOT to syllables with increasingly longer VOT. Participants identified all consonants with VOT below 25 msec as being /b/ and all consonant with VOT above 25 msec as being /p/. There was no grey area in which both interpretations of the sound were equally likely, which suggests that the phonemes were perceived categorically. Supporting findings also come from discrimination experiments: when one consonant has a VOT below 25 msec and the other above, people perceive the two syllables as being different (/ba/ and /pa/ respectively), but they do not notice any differences in the acoustic signal when both syllables have a VOT below or above 25 msec (even when these sounds are physically further away from each other than two sounds that cross the 25 msec dividing line).

Evidence from infant speech perception studies suggests that infants also perceive phonemes categorically (Eimas et al., 1971): one- and four-month old infants were presented with multiple syllables from the continuum of /ba/ to /pa/ sounds described above. As long as the syllables all came from above or below the 25 msec line, the infants showed no change in behavior (measured by their amount of sucking), but when presented with a syllable crossing that line, the infants reacted differently. This suggests that infants, just like adults, perceive speech sounds as belonging to discrete categories. Dehaene-Lambertz and Gliga (2004) also showed that the same neural systems are activated for both infants and adults when performing this task.

Importantly, languages differ in their phoneme inventories; for example English distinguishes /r/ from /l/ while Japanese does not, and children have to learn which categories to use. Experimental evidence suggests that infants can discriminate both native and nonnative speech sound differences up to 8 months of age, but have difficulty discriminating acoustically similar nonnative contrasts by 10-12 months of age (Werker and Hensch, 2015). These findings suggest that by their first birthday, they have learned to focus only on those contrasts that are relevant for their native language and to neglect those which are not. Psycholinguistic theories assume that children learn the categories of their native language by keeping track of the frequency distribution of acoustic sounds in their input. The forms around peaks in this distribution are then perceived as being a distinct category. Recent computational models showed that infant-directed speech contains sufficiently clear peaks for such a distributional learning mechanism to succeed and also that top-down processes like semantic knowledge and visual information play a role in phonetic category learning (ter Schure et al., 2016).

From the machine learning perspective categorical perception corresponds to the notion of learning invariances to certain properties of the input. With the experiments in Section 4 we attempt to gain some insight into this issue.

## 2.2 Computational models

There is a sizeable body of work on using recurrent neural (and other) networks to detect phonemes or phonetic features as a subcomponent of an ASR system. King and Taylor (2000) train recurrent neural networks to extract phonological features from framewise cepstral representation of speech in the TIMIT speaker-independent database. Frankel et al. (2007) introduce a dynamic Bayesian network for articulatory (phonetic) feature recognition as a component of an ASR system. Siniscalchi et al. (2013) show that a multilayer perceptron can successfully classify phonological features and contribute to the accuracy of a downstream ASR system.

Mohamed et al. (2012) use a Deep Belief Network (DBN) for acoustic modeling and phone recognition on human speech. They analyze the impact of the number of layers on phone recognition error rate, and visualize the MFCC vectors as well as the learned activation vectors of the hidden layers of the model. They show that the representations learned by the model are more speaker-invariant than the MFCC features.

These works directly supervise the networks to recognize phonological information. Another supervised but multimodal approach is taken by Sun (2016), which uses grounded speech for improving a supervised model of transcribing utterances from spoken description of images. We on the other hand are more interested in understanding how the phonological level of representation emerges from weak supervision via correlated signal from the visual modality.

There are some existing models which learn language representations from sensory input in such a weakly supervised fashion. For example Roy and Pentland (2002) use spoken utterances paired with images of objects, and search for segments of speech that reliably co-occur with visual shapes. Yu and Ballard (2004) use a similar approach but also include non-verbal cues such as gaze and gesture into the input for unsupervised learning of words and their visual meaning. These language learning models use rich input signals, but are very limited in scale and variation.

A separate line of research has used neural networks for modeling phonology from a (neuro)-cognitive perspective. Burgess and Hitch (1999) implement a connectionist model of the so-called phonological loop, i.e. the posited working memory which makes phonological forms available for recall (Baddeley and Hitch, 1974). Gasser and Lee (1989) show that Simple Recurrent Networks are capable of acquiring phonological constraints such as vowel harmony or phonological alterations at morpheme boundaries. Touretzky and Wheeler (1989) present a connectionist architecture which performs multiple simultaneous insertion, deletion, and mutation operations on sequences of phonemes. In this body of work the input to the network is at the level of phonemes or phonetic features, not acoustic features, and it is thus more concerned with the rules governing phonology and does not address how representations of phonemes arise from exposure to speech in the first place. Moreover, the early connectionist work deals with constrained, toy datasets. Current neural network architectures and hardware enable us to use much more realistic inputs with the potential to lead to qualitatively different results.

## 3 Model

As our model of language acquisition from grounded speech signal we adopt the Recurrent Highway Network-based model of Chrupała et al. (2017a). This model has two desirable properties: firstly, thanks to the analyses carried in that work, we understand roughly how the hidden layers differ in terms of the level of linguistic representation they encode. Secondly, the model is trained on clean synthetic speech which makes it appropri-ate to use for the controlled experiments in Section 5.2. We refer the reader to Chrupała et al. (2017a) for a detailed description of the model architecture. Here we give a brief overview.

The model exploits correlations between two modalities, i.e. speech and vision, as a source of weak supervision for learning to understand speech; in other words it implements language acquisition from the speech signal grounded in visual perception. The architecture is a bi-modal network whose learning objective is to project spoken utterances and images to a joint semantic space, such that corresponding pairs $(u, i)$ (i.e. an utterance and the image it describes) are close in this space, while unrelated pairs are far away, by a margin $\alpha$:

$$\sum_{u,i} \left( \sum_{u'} \max[0, \alpha + d(u,i) - d(u',i)] + \sum_{i'} \max[0, \alpha + d(u,i) - d(u,i')] \right) \quad (1)$$

where $d(u, i)$ is the cosine distance between the encoded utterance $u$ and encoded image $i$.

The image encoder part of the model uses image vectors from a pretrained object classification model, VGG-16 (Simonyan and Zisserman, 2014), and uses a linear transform to directly project these to the joint space. The utterance encoder takes Mel-frequency Cepstral Coefficients (MFCC) as input, and transforms it successively according to:

$$\mathrm{enc}_u(\mathbf{u}) = \mathrm{unit}(\mathrm{Attn}(\mathrm{RHN}_{k,L}(\mathrm{Conv}_{s,d,z}(\mathbf{u})))) \quad (2)$$

The first layer $\mathrm{Conv}_{s,d,z}$ is a one-dimensional convolution of size $s$ which subsamples the input with stride $z$, and projects it to $d$ dimensions. It is followed by $\mathrm{RHN}_{k,L}$ which consists of $k$ residualized recurrent layers. Specifically these are Recurrent Highway Network layers (Zilly et al., 2016), which are closely related to GRU networks, with the crucial difference that they increase the depth of the transform between timesteps; this is the recurrence depth $L$. The output of the final recurrent layer is passed through an attention-like lookback operator $\mathrm{Attn}$ which takes a weighted average of the activations across time steps. Finally, both utterance and image projections are L2-normalized. See Section 4.1 for details of the model configuration.

| | |
|---|---|
| Vowels | i ɪ ʊ u |
| | e ɛ ə ɚ ɔɪ ɔ o |
| | aɪ æ ʌ ɑ aʊ |
| Approximants | j ɹ l w |
| Nasals | m n ŋ |
| Plosives | p b t d k ɡ |
| Fricatives | f v θ ð s z ʃ ʒ h |
| Affricates | tʃ dʒ |

Table 1: Phonemes of General American English.

## 4 Experimental data and setup

The phoneme representations in each layer are calculated as the activations averaged over the duration of the phoneme occurrence in the input. The average input vectors are similarly calculated as the MFCC vectors averaged over the time course of the articulation of the phoneme occurrence. When we need to represent a phoneme type we do so by averaging the vectors of all its occurrences in the validation set. Table 1 shows the phoneme inventory we work with; this is also the inventory used by Gentle/Kaldi (see Section 4.3).

### 4.1 Model settings

We use the pre-trained version of the COCO Speech model, implemented in Theano (Bastien et al., 2012), provided by Chrupała et al. (2017a).[1] The details of the model configuration are as follows: convolutional layer with length 6, size 64, stride 3, 5 Recurrent Highway Network layers with 512 dimensions and 2 microsteps, attention Multi-Layer Perceptron with 512 hidden units, Adam optimizer, initial learning rate 0.0002. The 4096-dimensional image feature vectors come from the final fully connect layer of VGG-16 (Simonyan and Zisserman, 2014) pretrained on Imagenet (Russakovsky et al., 2014), and are averages of feature vectors for ten crops of each image. The total number of learnable parameters is 9,784,193. Table 2 sketches the architecture of the utterance encoder part of the model.

### 4.2 Synthetically Spoken COCO

The Speech COCO model was trained on the Synthetically Spoken COCO dataset (Chrupała et al., 2017b), which is a version of the MS COCO

---
[1] Code, data and pretrained models available from https://github.com/gchrupala/visually-grounded-speech.

| |
|---|
| Attention: size 512 |
| Recurrent 5: size 512 |
| Recurrent 4: size 512 |
| Recurrent 3: size 512 |
| Recurrent 2: size 512 |
| Recurrent 1: size 512 |
| Convolutional: size 64, length 6, stride 3 |
| Input MFCC: size 13 |

Table 2: COCO Speech utterance encoder architecture.

dataset (Lin et al., 2014) where speech was synthesized for the original image descriptions, using high-quality speech synthesis provided by gTTS.[2]

### 4.3 Forced alignment

We aligned the speech signal to the corresponding phonemic transcription with the Gentle toolkit,[3] which in turn is based on Kaldi (Povey et al., 2011). It uses a speech recognition model for English to transcribe the input audio signal, and then finds the optimal alignment of the transcription to the signal. This fails for a small number of utterances, which we remove from the data. In the next step we extract MFCC features from the audio signal and pass them through the COCO Speech utterance encoder, and record the activations for the convolutional layer as well as all the recurrent layers. For each utterance the representations (i.e. MFCC features and activations) are stored in a $t_r \times D_r$ matrix, where $t_r$ and $D_r$ are the number of times steps and the dimensionality, respectively, for each representation $r$. Given the alignment of each phoneme token to the underlying audio, we then infer the slice of the representation matrix corresponding to it.

## 5 Experiments

In this section we report on four experiments which we designed to elucidate to what extent information about phonology is represented in the activations of the layers of the COCO Speech model. In Section 5.1 we quantify how easy it is to decode phoneme identity from activations. In Section 5.2 we determine phoneme discriminability in a controlled task with minimal pair stimuli. Section 5.3 shows how the phoneme inventory is organized in the activation space of the model. Finally, in Section 5.4 we tackle the general issue of the representation of phonological form versus meaning with the controlled task of synonym discrimination.

### 5.1 Phoneme decoding

In this section we quantify to what extent phoneme identity can be decoded from the input MFCC features as compared to the representations extracted from the COCO speech. As explained in Section 4.3, we use phonemic transcriptions aligned to the corresponding audio in order to segment the signal into chunks corresponding to individual phonemes.

We take a sample of 5000 utterances from the validation set of Synthetically Spoken COCO, and extract the force-aligned representations from the Speech COCO model. We split this data into $\frac{2}{3}$ training and $\frac{1}{3}$ heldout portions, and use supervised classification in order to quantify the recoverability of phoneme identities from the representations. Each phoneme slice is averaged over time, so that it becomes a $D_r$-dimensional vector. For each representation we then train $L2$-penalized logistic regression (with the fixed penalty weight 1.0) on the training data and measure classification error rate on the heldout portion.

Figure 1 shows the results. As can be seen from this plot, phoneme recoverability is poor for the representations based on MFCC and the convolutional layer activations, but improves markedly for the recurrent layers. Phonemes are easiest recovered from the activations at recurrent layers 1 and 2, and the accuracy decreases thereafter. This suggests that the bottom recurrent layers of the model specialize in recognizing this type of low-level phonological information. It is notable however that even the last recurrent layer encodes phoneme identity to a substantial degree.

The MFCC features do much better than majority baseline (89% error rate) but poorly reltive to the the recurrent layers. Averaging across phoneme durations may be hurting performance, but interestingly, the network can overcome this and form more robust phoneme representations in the activation patterns.

### 5.2 Phoneme discrimination

Schatz et al. (2013) propose a framework for evaluating speech features learned in an unsupervised setup that does not depend on phonetically labeled data. They propose a set of tasks called Minimal-Pair ABX tasks that allow to make linguistically precise comparisons between syllable pairs that only differ by one phoneme. They use variants of this task to study phoneme discrimination across talkers and phonetic contexts as well as talker discrimination across phonemes.

---
[2]Available at https://github.com/pndurette/gTTS.
[3]Available at https://github.com/lowerquality/gentle.

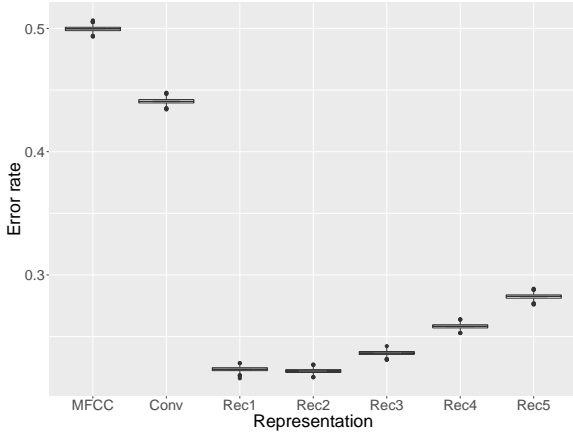

Figure 1: Accuracy of phoneme decoding with input MFCC features and COCO Speech model activations. The boxplot shows error rates bootstrapped with 1000 resamples.

Table 3: Accuracy of choosing the correct target in an ABX task using different representations.

| Representation | Accuracy |
|---|---|
| MFCC | 0.72 |
| Convolutional | 0.73 |
| Recurrent 1 | 0.83 |
| Recurrent 2 | 0.84 |
| Recurrent 3 | 0.80 |
| Recurrent 4 | 0.77 |
| Recurrent 5 | 0.75 |
| Embeddings | 0.67 |

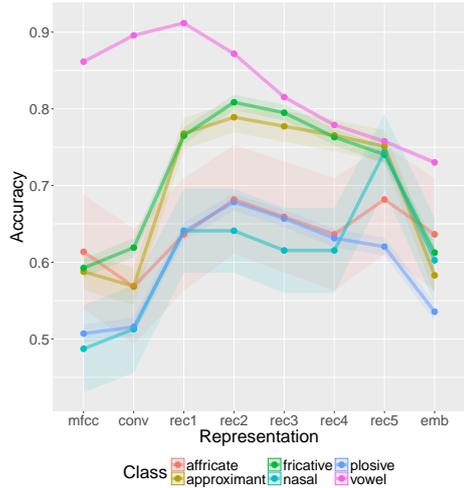

Figure 2: Accuracies for the ABX CV task for the cases where the target and the distractor belong to the same phoneme class. Shaded area extends $\pm 1$ standard error from the mean.

Here we evaluate the COCO Speech model on the *Phoneme across Context* (PaC) task of Schatz et al. (2013). This task consists of presenting a series of equal-length tuples $(A, B, X)$ to the model, where $A$ and $B$ differ by one phoneme (either a vowel or a consonant), as do $B$ and $X$, but $A$ and $X$ are not minimal pairs. For example, in the tuple (*be* /bi/, *me* /mi/, *my* /maɪ/), the task is to identify which of the two syllables /bi/ or /mi/ is closer to /maɪ/. The goal is to measure context invariance in phoneme discrimination by evaluating how often the model recognizes $X$ as the syllable closer to $B$ than to $A$.

We used a list of all attested consonant-vowel (CV) syllables of American English according to the syllabification method described in Gorman (2013). We excluded the ones which could not be unambiguously represented using English spelling for input to the TTS system (e.g. /baʊ/). We then compiled a list of all possible $(A, B, X)$ tuples from this list where $(A, B)$ and $(B, X)$ are minimal pairs, but $(A, X)$ are not. This resulted in 34,288 tuples in total. For each tuple, we measure $\text{sign}(\text{dist}(A, X) - \text{dist}(B, X))$, where $\text{dist}(i, j)$ is the euclidean distance between the vector representations of syllables $i$ and $j$. These representations are either the audio feature vectors or the layer activation vectors. A positive value for a tuple means that the model has correctly discriminated the phonemes that are shared or different across the syllables.

Table 3 shows the discrimination accuracy in this task using various representations. The pattern is similar to what we observed in the phoneme identification task: best accuracy is achieved using representation vectors from recurrent layers 1 and 2, and it drops as we move further up in the model. The accuracy is lowest when final embedding features are used for this task.

However, the PaC task is most meaningful and challenging where the target and the distractor phonemes belong to the same phoneme class. Fig-ure 2 shows the accuracies for this subset of cases, broken down by class. As can be seen, the model can discriminate between phonemes with high accuracy across all the layers, and the layer activations are more informative for this task than the MFCC features. Again, most phoneme classes seem to be represented more accurately in the lower layers (1–3), and the performance of the model in this task drops as we move towards higher hidden layers. There are also clear differences in the pattern of discriminability for the phoneme classes. The vowels are especially easy to tell apart, but accuracy on vowels drops most acutely in the higher layers. Meanwhile the accuracy on fricatives and approximants starts low, but improves rapidly and peaks around recurrent layer 2. The somewhat erratic pattern for nasals and affricates is most likely due to small sample size for these classes, as evident from the wide standard error.

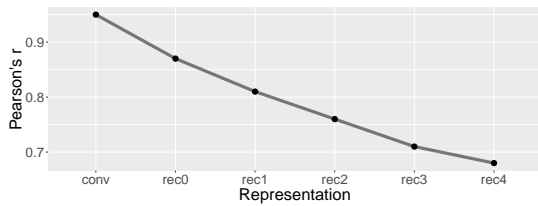

Figure 3: Pearson's correlation coefficients $r$ between the distance matrix of MFCCs and distance matrices on activation vectors.

### 5.3 Organization of phonemes

In this section we take a closer look at the underlying organization of phonemes in the model. Our experiment is inspired by Khalighinejad et al. (2017) who study how the speech signal is represented in the brain at different stages of the auditory pathway by collecting and analyzing electroencephalography responses from participants listening to continuous speech, and show that brain responses to different phoneme categories turn out to be organized by phonetic features.

We carry out an analogous experiment by analyzing the hidden layer activations of our model in response to each phoneme in the input. First, we generated a distance matrix for every pair of phonemes by calculating the Euclidean distance between the phoneme pair's activation vectors for each layer separately, as well as a distance matrix for all phoneme pairs based on their MFCC features. Similar to what Khalighinejad et al. (2017) report, we observe that the phoneme activations on all layers significantly correlate with the phoneme representations in the speech signal, and these correlations are strongest for the lower layers of the model. Figure 3 shows the results.

We then performed agglomerative hierarchical clustering on phoneme type MFCC and activation vectors, using Euclidean distance as the distance metric and the Ward linkage criterion (Ward Jr, 1963). Figure 5 shows the clustering results for the activation vectors on the first hidden layer. The leaf nodes are color-coded according to phoneme classes as specified in Table 1. There is substantial degree of matching between the classes and the structure of the hierarchy, but also some mixing between rounded back vowels and voiced plosives /b/ and /g/, which share articulatory features such as lip movement or tongue position.

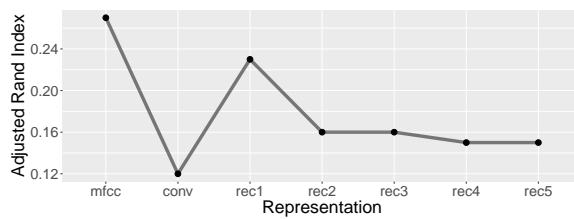

Figure 4: Adjusted Rand Index for the comparison of the phoneme type hierarchy induced from representations against phoneme classes.

We measured the adjusted Rand Index for the match between the hierarchy induced from each representation against phoneme classes, which were obtained by cutting the tree to divide the cluster into the same number of classes as there are phoneme classes. There is a notable drop between the match from MFCC to the activation of the convolutional layer. We suspect this may be explained by the loss of information caused by averaging over phoneme instances combined with the lower temporal resolution of the activations compared to MFCC. The match improves markedly at recurrent layer 1.

### 5.4 Synonym discrimination

Next we simulate the task of distinguishing between pairs of synonyms, i.e. words with different acoustic forms but the same meaning. With a representation encoding phonological form, our expectation is that the task would be easy; in contrast, with a representation which is invariant to phonological form in order to encode meaning, the task would be hard.

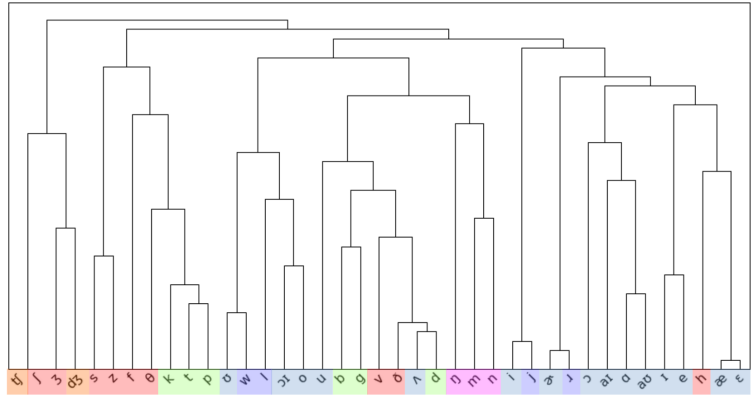

Figure 5: Hierarchical clustering of phoneme activation vectors on the first hidden layer.

We generate a list of synonyms for each noun, verb and adjective in the validation data using Wordnet (Miller, 1995) synset membership as a criterion. Out of these generated word pairs, we select synonyms for the experiment based on the following criteria:

- both forms clearly are synonyms in the sense that one word can be replaced by the other without changing the meaning of a sentence,
- both forms appear more than 20 times in the validation data,
- the words differ clearly in form (i.e. they are not simply variant spellings like *donut/doughnut, grey/gray*),
- the more frequent form constitutes less than 95% of the occurrences.

This gives us 2 verb, 2 adjective and 21 noun pairs.

For each synonym pair, we select the sentences in the validation set in which one of the two forms appears. We use the POS-tagging feature of NLTK (Bird, 2006) to ensure that only those sentences are selected in which the word appears in the correct word category (e.g. *play* and *show* are synonyms when used as nouns, but not when used as verbs). We then generate spoken utterances in which the original word is replaced by its synonym, resulting in the same amount of utterances for both words of each synonym pair.

For each pair we generate a binary classification task using the MFCC features, the average activations in the convolutional layer, the average unit activations per recurrent layer, and the sentence embeddings as input features. For every type of input, we run 10-fold cross validation using Logistic Regression to predict which of the two words the utterance contains. We used an average of 672 (minimum 96; maximum 2282) utterances for training the classifiers.

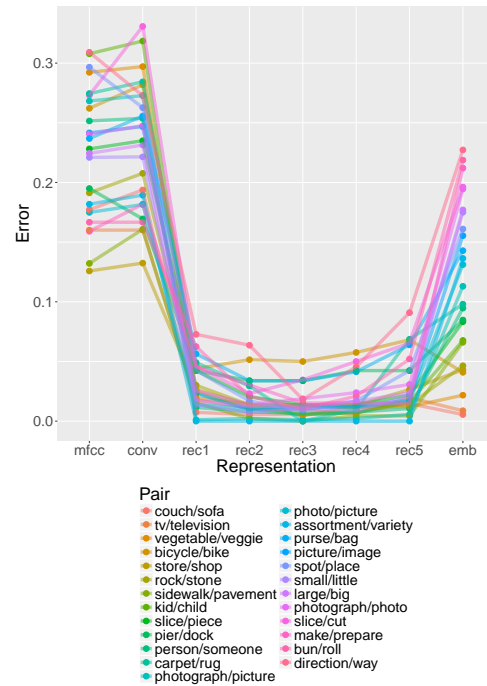

Figure 6: Synonym discrimination error rates, per representation and synonym pair.

Figure 6 shows the error rate in this classification task for each layer and each synonym pair. Recurrent layer activations are more informative for this task than MFCC features or activations of the convolutional layer. Across all the recurrent layers the error rate is small, showing that some form of phonological information is present throughout this part of the model. However, sentence embeddings give relatively high error rates suggesting that the attention layer acts to focus on semantic information and to filter out much of phonological form.

## 6 Discussion

Understanding distributed representations learned by neural networks is important but has the reputation of being hard or even impossible. In this work we focus on making progress on this problem for a particular domain: representations of phonology in a multilayer recurrent neural network trained on grounded speech signal. We believe it is important to carry out multiple analyses using diverse methodology: any single experiment may be misleading as it depends on analytical choices such as the type of supervised model used for decoding, the algorithm used for clustering, or the similarity metric for representational similarity analysis. To the extent that more than one experiment points to the same conclusion our confidence in the reliability of the insights gained will be increased.

Earlier work (Chrupała et al., 2017a) shows that encoding of semantics in our RNN model of grounded speech becomes stronger in higher layers, while encoding of form becomes weaker. The main high-level results of our study confirm this pattern by showing that the representation of phonological knowledge is most accurate in the lower layers of the model. This general pattern is to be expected as the objective of the utterance encoder is to transform the input acoustic features in such a way that it can be matched to its counterpart in a completely separate modality. Many of the details of how this happens, however, are far from obvious: perhaps most surprisingly we found that a large amount of phonological information is still available up to the top recurrent layer. Evidence for this pattern emerges from the phoneme decoding task, the ABX task and the synonym discrimination task. The last one also shows that the attention layer filters out and significantly attenuates encoding of phonology and makes the utterance embeddings much more invariant to synonymy.

Our model is trained on synthetic speech, which is easier to process than natural human-generated speech. While small-scale databases of natural speech and image are available (e.g. the Flickr8k Audio Caption Corpus, Harwath and Glass, 2015), they are not large enough to reliably train models such as ours. In future we would like to collect more data and apply our methodology to grounded human speech and investigate whether context and speaker-invariant phoneme representations can be learned from natural, noisy input. We would also like to make comparisons to the results that emerge from similar analyses applied to neuroimaging data.